\documentclass[10pt,twocolumn,letterpaper]{article}

\usepackage{cvpr}              % To produce the CAMERA-READY version
\usepackage{graphicx}
\usepackage{amsmath}
\usepackage{amssymb}
\usepackage{booktabs}
\usepackage[accsupp]{axessibility}

\DeclareMathOperator*{\argmin}{arg\,min}
\usepackage[pagebackref,breaklinks,colorlinks]{hyperref}

\usepackage[capitalize]{cleveref}
\crefname{section}{Sec.}{Secs.}
\Crefname{section}{Section}{Sections}
\Crefname{table}{Table}{Tables}
\crefname{table}{Tab.}{Tabs.}

\def\cvprPaperID{6} % *** Enter the CVPR Paper ID here
\def\confName{CVPR}
\def\confYear{2022}

\newcommand{\longrightleftarrows}[1]{\mathrel{\substack{\xrightarrow{#1} \\[-0.9ex] \xleftarrow{#1}}}}
\begin{document}
	
	\title{Generative Flows as a General Purpose Solution for Inverse Problems}
	
	\author{José A. Chávez\\
		Universidad Católica San Pablo\\
		Arequipa, Perú\\
		{\tt\small jose.chavez.alvarez@ucsp.edu.pe}
	%\and
}
\maketitle

%%%%%%%%% ABSTRACT
\begin{abstract}
	Due to the success of generative flows to model data distributions, they have been explored in inverse problems. Given a pre-trained generative flow, previous work proposed to minimize the 2-norm of the latent variables as a regularization term. The intuition behind it was to ensure high likelihood latent variables that produce the closest restoration. However, high-likelihood latent variables may generate unrealistic samples as we show in our experiments. We therefore propose a solver to directly produce high-likelihood reconstructions. We hypothesize that our approach could make generative flows a general purpose solver for inverse problems. Furthermore, we propose $1\times 1$ coupling functions to introduce permutations in a generative flow. It has the advantage that its inverse does not require to be calculated in the generation process. Finally, we evaluate our method for denoising, deblurring, inpainting, and colorization. We observe a compelling improvement of our method over prior works.
\end{abstract}

%%%%%%%%% BODY TEXT
\section{Introduction}
\label{sec:intro}

With the success of generative models in the generation of synthetic data \cite{kingma2014vae,goodfellow2014gan}, prior works \cite{bora2017compressed,ardizzone2018analyzing,asim2020invertible} explored them for solving inverse problems in image processing. In this approach, a generative model can be pre-trained with some data such as medical images, faces, animals, and more \cite{YI2019101552, karras2021alias, wu2019logan}. Then, this trained model is used to find the best restoration for an inverse problem.

In an inverse problem we aim to recover $x^{*}\in\mathbb{R}^{n}$ from the possible-noisy linear measurements $y=Ax^{*} +\eta$, where $A\in\mathbb{R}^{m\times n}$ is the matrix measurement and $\eta\in\mathbb{R}^{m}$ is the noise. For instance, $y$ would be a noisy or blurred image, and $x^{*}$ would be the restoration of this image. So given a generative model $G:z\rightarrow x$, researchers \cite{bora2017compressed} proposed the following objective function

%Image denoising represents a critic problem in several tasks such as semantic segmentation, classification and object tracking. 
\begin{equation}%
	\label{eq:1}
	\hat{z} = \argmin_{z\in \mathbb{R}^k}\| AG(z)-y\|^{2} + \gamma\cdot\| z\|^{2},
\end{equation}

\noindent  where $z\in\mathbb{R}^{k}$ and $x\in\mathbb{R}^{n}$. And the most typical choice for $\| \cdot \|$ is the 2-norm~\cite{bora2017compressed, asim2020invertible}. Note that the network $G$ was trained with similar samples than $x^{*}$, thus the intuition is to find $\hat{z}$ such that the synthetic image $\hat{x}=G(\hat{z})$ is close to $x^{*}$.
%the latent variable $\hat{z}$ that minimize the Euclidean Norm $\| AG(z)-y\|^{2}$. 
%\noindent where both $\|\cdot\|$ are the 2-norm. 

% In contrast to VAEs \citep{kingma2014vae} and GANs \citep{goodfellow2014gan}, 

Generative Adversarial Networks (GANs) have shown great success in image synthesis \cite{karras2021alias, sauer2022stylegan}; however they are not invertible. Experiments in image manipulation with GANs \cite{abdal2021styeflow} have shown that \textit{projected images} \cite{zhu2016manipulation} are not identical to the original ones. Thus, even if we set the identity matrix as $A$, and $\eta=0$ in Equation~\ref{eq:1}; we will probably never achieve the zero-error reconstruction ($x^{*}=\hat{x}$). On the other hand, Variational Auto-Encoders (VAEs) perform a stochastic mapping from samples to latent variables. Although VAEs have an encoder, we can infer only approximately the corresponding latent variable for a given sample.

\begin{figure*}
	\centering
	%\hspace{-1.5mm}
	\rotatebox{90}{\hspace{1.8mm}$0.0$}
	\includegraphics[height=1.1cm,trim={0 0 0 0},clip]{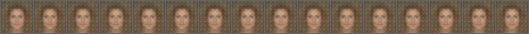}\\
	\rotatebox{90}{\hspace{1.85mm}$0.2$}
	\includegraphics[height=1.1cm,trim={0 0 0 0},clip]{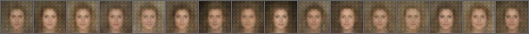}\\
	\rotatebox{90}{\hspace{1.90mm}$0.4$}
	\includegraphics[height=1.1cm,trim={0 0 0 0},clip]{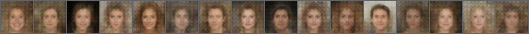}\\
	\rotatebox{90}{\hspace{1.95mm}$0.6$}
	\includegraphics[height=1.1cm,trim={0 0 0 0},clip]{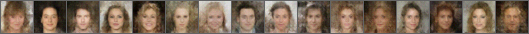}\\
	\rotatebox{90}{\hspace{2.0mm}$0.8$}
	\includegraphics[height=1.1cm,trim={0 0 0 0},clip]{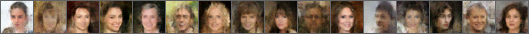}\\
	\rotatebox{90}{\hspace{2.0mm}$1.0$}
	\includegraphics[height=1.1cm,trim={0 0 0 0},clip]{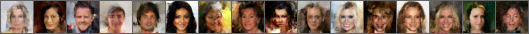}\\
	\rotatebox{90}{\hspace{2.04mm}$1.2$}
	\includegraphics[height=1.1cm,trim={0 0 0 0},clip]{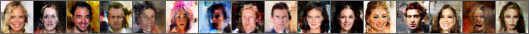}\\
	\rotatebox{90}{\hspace{2.08mm}$1.4$}
	\includegraphics[height=1.1cm,trim={0 0 0 0},clip]{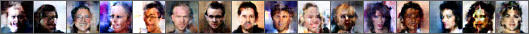}\\
	\rotatebox{90}{\hspace{2.12mm}$1.6$}
	\includegraphics[height=1.1cm,trim={0 0 0 0},clip]{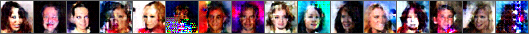}\\
	\rotatebox{90}{\hspace{2.16mm}$1.8$}
	\includegraphics[height=1.1cm,trim={0 0 0 0},clip]{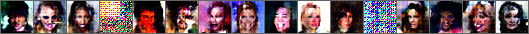}\\
	\rotatebox{90}{\hspace{2.2mm}$2.0$}
	\includegraphics[height=1.1cm,trim={0 0 0 0},clip]{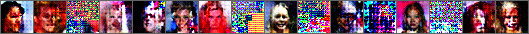}
	\caption{Samples from our generative flow trained on the CelebA dataset. We generate images from an isotropic Gaussian distribution $\mathcal{N}(\mathbf{0},\sigma^2 \mathbf{I})$. Each row represents a batch of $16$ samples by setting a specific value for $\sigma$. Note that $\sigma=0$ represents $z=\mathbf{0}$.}
	\label{fig:samples}
\end{figure*}

Generative flows \cite{dinh2014nice} have a particular advantage among generative models; they are invertible networks. Hence, these models can perform the re-generation of any image---there exists a latent variable for any image $x^{*}$ in an inverse problem, and is unique. A sequence of bijective functions allows the exact reconstruction of a sample with its corresponding latent variable. Therefore, zero-error reconstruction is possible and easy to achieve. Authors in~\cite{ardizzone2018analyzing} proposed to solve a unique inverse problem with generative flows. In this approach, the aim is to encode the corresponding measurements $y$ from a given image $x^{*}$. Then, once the model is trained with synthetic pairs of $\{x^{*},y\}$, it may estimate the corresponding original image from some measurements $y$. Note that this method recovers the original image from its measurements once it is trained on a specific inverse problem. One must train different pairs of $\{x^{*},y\}$ for each task.

In contrast, GlowIP \cite{asim2020invertible} explored pre-trained generative flows for compressed sensing, denoising and inpainting. In GlowIP, the Glow architecture~\cite{kingma2018glow} is trained on the CelebA dataset~\cite{liu2015deep}, thus encoding a latent variable for each image as $G^{-1}:x\rightarrow z$. Once the model is trained, GlowIP performs Equation~\ref{eq:1} to solve an inverse problem. This model does not require the knowledge of any specific inverse problem in the training process---it is only trained with images from CelebA.

Note that Equation~\ref{eq:1} has a regularization term $\| z\|^{2}$. As authors suggest in \cite{bora2017compressed,asim2020invertible}, the intuition behind it is to encourage high likelihood latent variables, because $G$ is typically trained  with the latent distribution $\mathcal{N}(\mathbf{0},\mathbf{I})$. However, a high log-likelihood in the latent representation may generate a low likelihood sample $x$. Indeed, the sample $x_0=G(\mathbf{0})$ is a blurry and unlikely image in a generative flow (see Figure \ref{fig:samples}). Our hypothesis is that $z$ must produce realistic samples instead. Recently, researchers~\cite{whang2021solving} have attempted to generalize Equation~\ref{eq:1}, they proposed a maximum a posteriori (MAP) objective for image denoising as follows

\begin{equation}
	\label{eq:20}
	\begin{aligned}
		\hat{z} = \argmin_{z\in \mathbb{R}^k}\frac{1}{2\sigma^2}\| G(z)-y\|^{2} -\beta \cdot \log p_G(G(z)),
	\end{aligned}
\end{equation} 

\noindent where $\sqrt{\mathbb{E}[\|\eta\|^{2}]}=\sigma$, and $p_{G}(G(z))$ is the density estimation by $G$. However, one must be careful when using density estimates from a generative flow. In the training process, $G$ only sees samples from its corresponding dataset. Nevertheless, one may obtain the density estimates from out-of-distribution samples. In fact, a generative flow may assign a high likelihood to constant or random images~\cite{nalisnick2018deep}. Additionally, the estimation of $p_G(G(z))$ involves using the generative flow in the form of $G$ and then $G^{-1}$. As the authors suggest in \cite{whang2021solving}, the hyperparameter $\beta$ should be configured to control this problem. 

On the other hand, permutations represent a key process in generative flows to handle the expressiveness of bijective functions. The Glow architecture~\cite{kingma2018glow} requires invertible $1\times 1$ convolutions to permute the entries of hidden feature maps. Note that this architecture requires computing the inverse of the invertible $1\times 1$ convolutions in order to solve Equation~\ref{eq:1} or Equation~\ref{eq:20}. Invertible $1\times 1$ convolutions have shown impressive performance in the generation of synthetic high-resolution images. Nonetheless, the inverse of this layer has a significant cost in limited resources, which could be avoided.

In this work, we aim to explore pre-trained generative flows as a general purpose solution for inverse problems. The contributions are summarized in the following points:

\begin{itemize}
	\item A solver to produce realistic restorations in inverse problems through a pre-trained generative flow. Our approach demonstrates quantitive and qualitative improvement among the state-of-the-art methods in many inverse problems. 
	
	\item We introduce $1\times 1$ coupling functions to permute the channels of feature maps in a generative flow. This permutation does not require the computation of its inverse during the generation of samples.
\end{itemize}

Code to implement our solver and reproduce results is publicly available\footnote{The code is available at \url{https://github.com/jarchv/solverip.git}}.

%-------------------------------------------------------------------------

\section{Preliminaries}

In this section, we give an overview on the generative flow model as well as existing methods to introduce permutations. 

\subsection{Generative flows}
Generative flows can be trained by the maximum likelihood using the change of variable formula

\begin{equation}
	\label{eq:2}
	\begin{aligned}
		\log(p_{X}(x))=\log(p_{Z}(z)) + \log(|\det\mathbf{J}_{F}(x)|),
	\end{aligned}
\end{equation} 

\noindent where $F:x\rightarrow z$ is an invertible network that maps from a sample $x\sim p_{X}$ to the latent variable $z\sim p_{Z}$; and $\mathbf{J}_{F}(x)=\frac{\partial F(x)}{\partial x}$ is the Jacobian of $F$ evaluated at $x$. The network $F$ is composed of a sequence of bijections as follows

\begin{equation}
	\label{eq:10}
	x \underset{F^{-1}_1}{\overset{F_1}{\longrightleftarrows{\rule{0.7cm}{0cm}}}} h^{[1]} \underset{F^{-1}_2}{\overset{F_2}{\longrightleftarrows{\rule{0.7cm}{0cm}}}} h^{[2]} \cdots h^{[l-1]}\underset{F^{-1}_l}{\overset{F_l}{\longrightleftarrows{\rule{0.7cm}{0cm}}}} h^{[l]}=z,
\end{equation}

\noindent where $F$ must be a transformation such that its Jacobian determinant is computationally tractable \cite{dinh2014nice}. The idea is to choose transformations $F_i$ whose Jacobian $\frac{\partial h_{i}}{\partial h_{i-1}}$ represents a triangular matrix---the determinant of a triangular matrix is the product of its diagonal elements. Training $F$ in Equation~\ref{eq:2} with maximum likelihood is also known as \textit{Gaussianization}\cite{chen2000gauss}, where the dimensions in $z$ becomes independent and therefore

\begin{equation}
	\label{eq:11}
	p_{Z}(z) = \prod_{i=1}^{k}p_{Z_i}(z_i),
\end{equation} 

where $k$ is number of entries of $z$. Note that Equation~\ref{eq:11} simplifies the maximization of the right-hand side in Equation~\ref{eq:2}. Moreover, each bijection $F_i$ is also called \textit{coupling layer}\cite{dinh2014nice}, and it is a reversible triangular transformation where its determinant is easy to compute. 

\subsection{Permutations in generative flows}

A coupling layer leaves half of its input unchanged while the other is modified. This expressiveness of coupling layers caused by the half unchanged can be handled with permutations. In NICE~\cite{dinh2014nice}, a \texttt{swap} operation exchanges the halves of the coupling layer input, so in two consecutive coupling layers, the entire input is modified. The Glow model~\cite{kingma2018glow} implements learned invertible $1\times 1$ convolutions, which performs impressive results in the CelebA HQ dataset~\cite{karras2017progressive}. However, as the authors said the cost of computing its determinant is $\mathcal{O}(c^3)$, where $c$ is the number of input channels in the invertible $1\times 1$ convolution. Furthermore, its inverse has a cost of $\mathcal{O}(c^3)$. On the other hand, Glow resizes the feature maps with squeeze layers \cite{dinh2017realnvp}. Due to a squeeze layer should produce the same number of elements for a given input, it increments the number of channels $c$ that will be permuted by invertible $1\times 1$ convolutions. This process increases the number of parameters of invertible $1\times1$ convolution, and consequently the cost of computing its Jacobian determinant and its inverse.

The Glow architecture \cite{kingma2018glow} is used in GlowIP \cite{asim2020invertible} for inverse problems. Note the process of solving inverse problems in Equation~\ref{eq:1} involves using the inverse of the network. It means that GlowIP should compute the inverse of each invertible $1\times 1$ convolution in the sample generation. Moreover, training the Glow architecture to infer latent variables from images also involves computing the determinant of this permutation layer.

%------------------------------------------------------------------------

\section{Proposed approach}

In this section, we first describe our inverse problem solver to generate realist reconstructions, and then our proposed $1\times 1$ coupling functions to introduce permutations.

\subsection{Inverse problem solver}

Once we our generative flow $F_{\phi}$ is trained (parametrized by $\phi$) with the CelebA dataset\cite{liu2015deep}, we proceed to solve inverse problems. Let assume that we aim to recover $x^{*}\in\mathbb{R}^{n}$ from possible-noisy linear measurements $y=Ax^{*}+\eta$, where $A\in\mathbb{R}^{m\times n}$ and $\eta\in\mathbb\mathbb{R}^{m}$. Then, given a pre-trained generative flow $F_{\phi}:x\rightarrow z$ , where $x,z\in\mathbb{R}^{n}$, we may rewrite Equation \ref{eq:2} as 

\begin{align}
	\label{eq:8}
	\log(p_{\phi}(x))&=\log(p_{Z}(z)) - \log(|\det\mathbf{J}_{F_{\phi}^{-1}}(z)|),
\end{align}

%=\frac{\partial F_{\phi}^{-1}(z)}{\partial z}

\noindent where $\log(p_{\phi}(x))$ is the log-likelikehood. Note that we replace the term $|\det\mathbf{J}_{F_{\phi}}(x)|$ by $|\det\mathbf{J}_{F_{\phi}^{-1}}(z)|^{-1}$ in Equation~\ref{eq:2}. Now, $\mathbf{J}_{F_{\phi}^{-1}}(z)$ is the Jacobian of $F_{\phi}^{-1}$ evaluated at $z$.

Note that the maximization of $\log(p_{Z}(z))$ is equivalent to the minimization of $\|z\|^2$, which is proposed in \cite{bora2017compressed, asim2020invertible}. However, it does not ensure the maximization of the log-likelihood $\log(p_{\phi}(x))$---we must maximize the Jacobian term as well. The maximization of the right-hand side in Equation \ref{eq:8} directly ensures the generation of high log-likelihood samples $x$, which is what we pretend. Then, the objective function to find the latent variable that recovers $x^{*}$ from $y$ in an inverse problem is

\begin{align}
	\label{eq:9}
	\hat{z} &= \argmin_{z\in \mathbb{R}^k}\left\{\| AF^{-1}(z)-y\|_{1} + \alpha\cdot\mathcal{L}(z)\right\},\\
	\mathcal{L}(z) &= - \log(p_{Z}(z)) + \log(|\det\mathbf{J}_{F_{\phi}^{-1}}(z)|),
\end{align}

\noindent where $\alpha>0$ is a hyperparameter. We find empirically that our regularization term $\mathcal{L}(z)$ performs better with the 1-norm in Equation~\ref{eq:9}. Thus, while our solver generates realistic samples, it is also finding the optimal reconstruction of the inverse problem.

% The purpose of introducing the negative log-likelihood term into the main objective is to generate realistic images in the reconstruction process of inverse problems.

\subsection{$1\times 1$ Coupling functions}

\begin{figure}
	\centering
	\includegraphics[height=2.9cm, trim={2.2cm 0 0.8cm 0},clip]{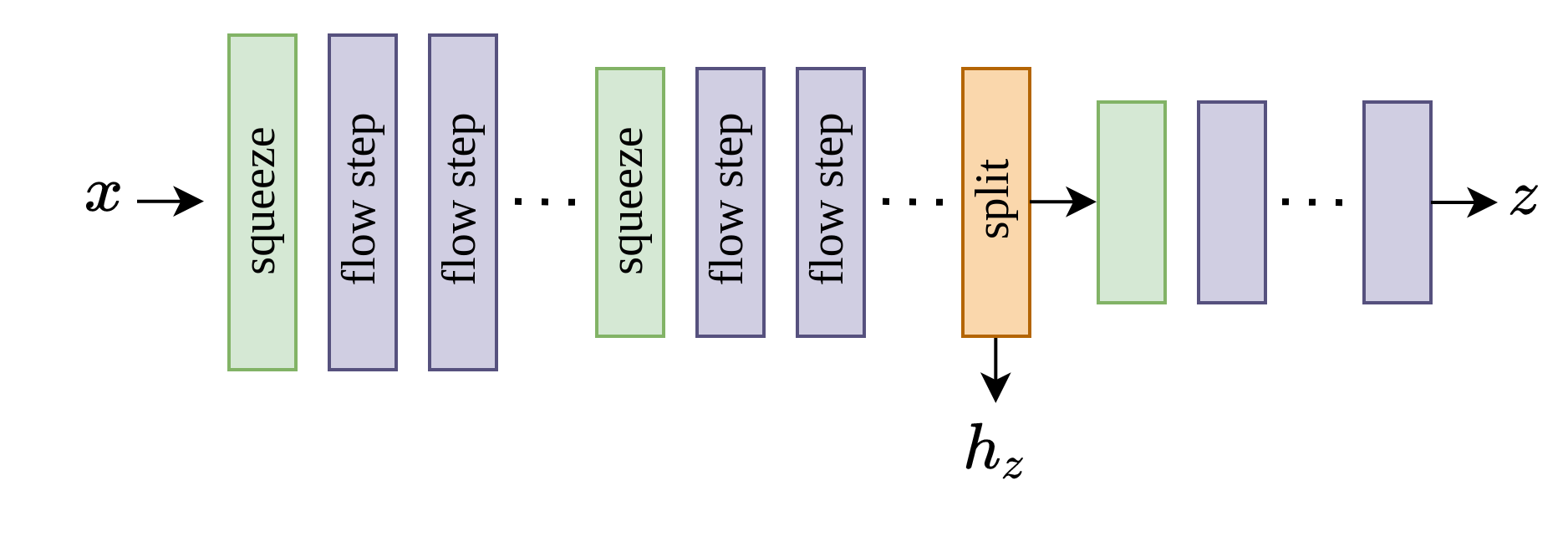}
	\caption{Our generative flow architecture. Each \textit{flow step} is composed by an activation normalization \cite{kingma2018glow} followed by our proposed coupling layer. These flow steps are combined with squeeze layers. We also split low-scale feature maps in two chunks, and force one of the halves ($h_{z}$) to directly follow $p_Z$. The other half continues its path until produce the $1\times 1$ latent variable $z$.}
	\label{fig:architecture}
\end{figure}

Figure \ref{fig:architecture} shows our generative flow architecture. A \textit{flow step} represents a sequence of an activation normalization~\cite{kingma2018glow} and a coupling layer. Inside a coupling layer, we implement $\texttt{swap}$ operations. It introduce permutations as follow

\begin{align}
	h_1, h_2 = \texttt{swap}(\texttt{split}(h)) \label{eq:3},
\end{align}

\noindent where $\texttt{split}$ separates the input tensor $h\in\mathbb{R}^{n}$ into two chunks $h_1\in\mathbb{R}^{n/2}$ and $h_2\in\mathbb{R}^{n/2}$. Note that a $\texttt{swap}$ operation has constant time complexity $\mathcal{O}(1)$. Then, we set a convolutional neural network $\texttt{CNN}$ as the \textit{coupling function}, which process the first chunk as

\begin{align}
	s, t = \texttt{CNN}(h_1) \label{eq:4},
\end{align}

\noindent where $s$ and $t$ have the dimension as $h_1$. Inspired on Glow~\cite{kingma2018glow}, we perform the following affine transformation on $h_2$

\begin{align}
	h'_2 &= h_2 \odot  \texttt{sigmoid}(s + 2) + t \label{eq:5},
\end{align}

\noindent where $\texttt{sigmoid}$ is the sigmoid function. Finally, we apply a $\texttt{concat}$ operation to obtain the output tensor $h'$, which has the same dimension as the input $h$

\begin{align}
	h' &= \texttt{concat}(\texttt{swap}(h_1, h'_2)) \label{eq:6}.
\end{align}

Note that if we apply a $\texttt{swap}$ operation in Equation \ref{eq:3}, then we should apply another one in Equation \ref{eq:6} to hold the order of the entries in $h$.% This operation is applied in alternating coupling layers; thus, every element of $h$ is modified in two consecutive coupling layers. 

An additional permutation is performed in our generative flow architecture. Our coupling function applies $1\times1$ convolutions on $1\times 1$ feature maps $h$. Hence, each $1\times1$ coupling function permutes all the elements of $1\times 1$ features maps. However, each of these coupling functions does not require the computation of its inverse during sample generation. Finally, in the process of training our generative flow, the Jacobian of $1\times1$ coupling function is not required to be computed.

% Finally, we use $512$ hidden channels in the network $\texttt{NN}$.

\begin{figure*}
	\centering
	\hspace{-1.5mm}
	\rotatebox{90}{\hspace{1.0mm} Target}
	\includegraphics[height=1.35cm,trim={15.95cm 0 0 0},clip]{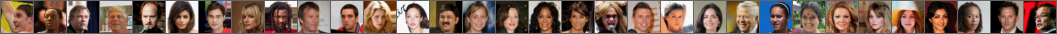}\\
	\hspace{-1.5mm}
	\rotatebox{90}{\hspace{1.4mm} Noisy}
	\includegraphics[height=1.35cm,trim={15.95cm 0 0 0},clip]{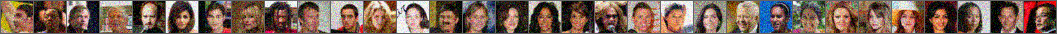}\\
	\rotatebox{90}{\hspace{0.4mm} CSGM}
	\includegraphics[height=1.35cm,trim={15.95cm 0 0 0},clip]{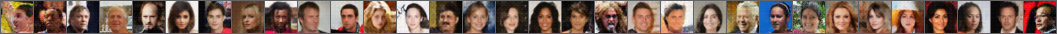}\\
	\rotatebox{90}{\hspace{0mm} GlowIP}
	\includegraphics[height=1.35cm,trim={15.95cm 0 0 0},clip]{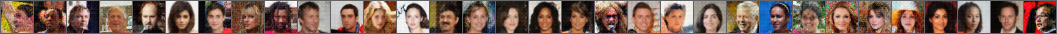}\\
	\rotatebox{90}{\hspace{0mm} MAP}
	\includegraphics[height=1.35cm,trim={15.95cm 0 0 0},clip]{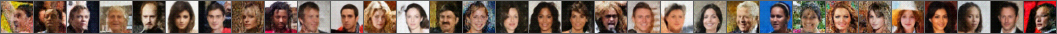}\\
	\rotatebox{90}{\hspace{0.4mm} BM3D}
	\includegraphics[height=1.35cm,trim={15.95cm 0 0 0},clip]{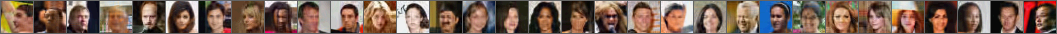}\\
	\rotatebox{90}{\hspace{1.5mm} Ours}
	\includegraphics[height=1.35cm,trim={15.95cm 0 0 0},clip]{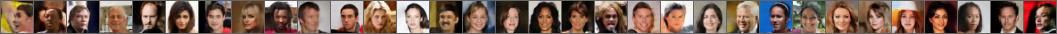}
	
	\caption{Image denoising on images from the CelebA dataset. The first row is the target image $x^{*}$ and the second row is the noisy image $y$, we aim to recover $x^{*}$ from $y$. From the 3rd-7th rows, we show results for CSGM, GlowIP, MAP, BM3D and our proposed solver. We synthetically create noisy images with a noise level of $\sqrt{\mathbb{E}[\|\eta\|^{2}]}=0.1$.}
	\label{fig:denoising}
\end{figure*}

\begin{figure*}
	\centering
	\hspace{-1.5mm}
	\rotatebox{90}{\hspace{1.8mm}Target}
	\includegraphics[height=1.35cm,trim={0 0 15.95cm 0},clip]{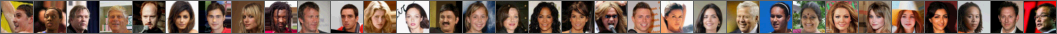}\\
	\hspace{-1.5mm}
	\rotatebox{90}{\hspace{1.8mm}Blurry}
	\includegraphics[height=1.35cm,trim={0 0 15.95cm 0},clip]{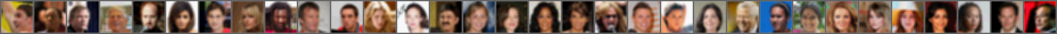}\\
	\rotatebox{90}{\hspace{1.2mm}CSGM}
	\includegraphics[height=1.35cm,trim={0 0 15.95cm 0},clip]{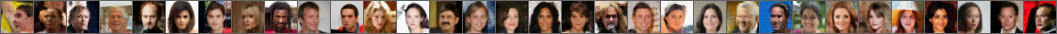}\\
	\rotatebox{90}{\hspace{1mm}GlowIP}
	\includegraphics[height=1.35cm,trim={0 0 15.95cm 0},clip]{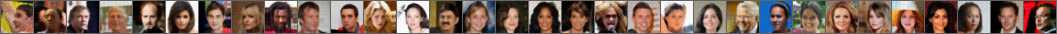}\\
	\rotatebox{90}{\hspace{2.8mm}Ours}
	\includegraphics[height=1.35cm,trim={0 0 15.95cm 0},clip]{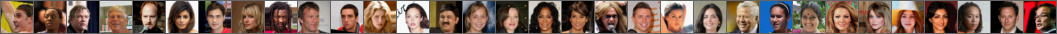}
	
	\caption{Image deblurring on images from the CelebA dataset. The first row is the target image $x^{*}$ and the second row is the blurry image $y$, we aim to recover $x^{*}$ from $y$. From the 3rd-5th rows, we show results for CSGM, GlowIP, and our proposed solver. We create synthetic blurry images by applying a $3\times3$ average pooling with stride $1$ on targets images.} 
	\label{fig:deblurring}
\end{figure*}

%------------------------------------------------------------------------
\section{Experiments}

\begin{figure*}
	\centering
	\hspace{-1.5mm}
	\rotatebox{90}{\hspace{1.8mm}Target}
	\includegraphics[height=1.35cm,trim={0cm 0 15.95cm 0},clip]{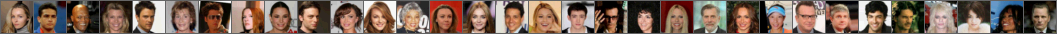}\\
	\rotatebox{90}{\hspace{0.8mm}Masked}
	\includegraphics[height=1.35cm,trim={0cm 0 15.95cm 0},clip]{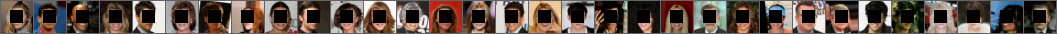}\\
	\rotatebox{90}{\hspace{1.2mm}CSGM}
	\includegraphics[height=1.35cm,trim={0cm 0 15.95cm 0},clip]{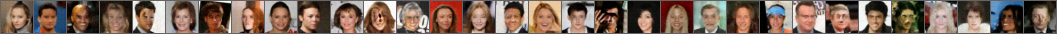}\\
	\rotatebox{90}{\hspace{1mm}GlowIP}
	\includegraphics[height=1.35cm,trim={0cm 0 15.95cm 0},clip]{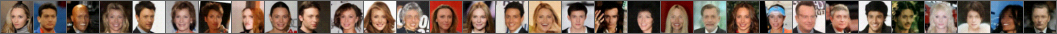}\\
	\rotatebox{90}{\hspace{2.8mm}Ours}
	\includegraphics[height=1.35cm,trim={0cm 0 15.95cm 0},clip]{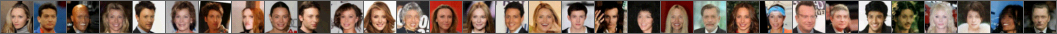}
	
	\caption{Image inpainting on images from the CelebA dataset. The first row is the target image $x^{*}$ and the second row is the masked image $y$, we aim to recover $x^{*}$ from $y$. From the 3rd-5th rows, we show results for CSGM, GlowIP, and our proposed solver. We crop a region of images to create the masked images.}
	\label{fig:inpainting}
\end{figure*}

This section illustrates the experimental details and presents the comparison of our proposed solver to other state-of-the-art methods. The performance evaluation is split into four parts: denoising, deblurring, inpainting, and colorization. Then, we show how realistic are samples with different standard deviations in the latent distribution. Finally, we conduct an ablation study to compare the time speed and memory usage when we use $1\times1$ coupling functions and invertible $1\times 1$ convolution.

\section{Training details and parameters settings}

First, we resize to $32\times32$ resolution all images of the CelebA dataset \cite{liu2015deep}. Then, we fix a generative flow with the training samples. The Adam optimizer~\cite{kingma2015adam} is used during the training with parameters $\alpha=0.0001$, $\beta_1=0.5$ and $\beta_2=0.999$. Inspired on \cite{dinh2017realnvp}, we force half of the low-scale feature maps to follow the latent distribution as we can see in Figure \ref{fig:architecture}. We perform \textit{squeeze layers}~\cite{dinh2017realnvp} to reduce and increase the scale of feature maps. Additionally, we set $L=5$ as the number of scales, and $K=2$ as the number of flow steps per scale. The value of $K$ is duplicated for scales $2\times2$ and $1\times1$---we use $14$ flow steps in total. We set the batch size as $32$ and stop the training process after $100$ epochs. Finally, we use $512$ hidden channels for each network $\texttt{CNN}$.

Once the model is trained, we optimize Equation \ref{eq:9} and initialize the latent variable at $z_0\sim\mathcal{N}(\mathbf{0},\sigma^2\mathbf{I})$ for $\sigma=0.1$. In this process, we set $\alpha=0.005$, $\beta_1=0.9$ and $\beta_2=0.999$. We configure the batch size as $32$, and after $1500$ iterations we return the restored image for the respective inverse problem. All experiments are made in an NVIDIA GeForce GTX 750 Ti. 

To compare with other methods such as CSGM \cite{bora2017compressed} and GlowIP \cite{asim2020invertible}, we use their initialization and objective function in our pre-trained generative flow. Note that CSGM and GlowIP perform the same objective function (Equation \ref{eq:1}), the unique difference is the initialization: random initialization for CSGM and zero initialization for GlowIP. We therefore test many values for their penalization hyperparameter $\gamma$. Each value of this hyperparameter in our experiments is the one that achieves the best performance for each inverse problem. Additionally, in image denoising, we also compare with BM3D~\cite{dabov2007image} and MAP~\cite{whang2021solving}. We use the PSNR and SSIM metrics to measure the quality of restorations on each inverse problem.

\subsection{Denoising}

For image denoising we use a noise level of $\sqrt{\mathbb{E}[\|\eta\|^{2}]}=0.1$. Figure \ref{fig:denoising} shows that CSGM ($\gamma=0.1$) smooths the noisy image but can not remove all distortions. These distortions are slightly more visible in GlowIP ($\gamma=0.1$) than CSGM. On the other hand, MAP~\cite{whang2021solving} has similar visual results to GlowIP. In MAP, we use the same initialization as ours, and $0.0015$ as the learning rate with $\beta=0.5$---we find empirically the best performance for this method using these settings. Our method outperforms BM3D in the PSNR and SSIM metrics as you can see in Table \ref{tab:measures}. Indeed, our restorations look sharper and more realistic than those from BM3D (see Figure \ref{fig:denoising}). Nonetheless, our method eliminates some details from the background in samples. In this case, we set $\alpha=0.05$ in Equation \ref{eq:9}. 

\subsection{Deblurring}

We create synthetic blurry images by applying an average pooling in the target image $x^{*}$, and use it as the measurements $y$. There is no significant difference between CSGM ($\gamma=0.01$) and GlowIP ($\gamma=0$) in image deblurring (see Figure \ref{fig:deblurring}). Both produce visual artifacts that are exacerbated if we increase their penalization term---CSGM and GlowIP mitigate visual artifacts with $\gamma=0.01$ and $\gamma=0$ respectively. As you can see in Table \ref{tab:measures}, our method outperforms GlowIP and CSGM in the PSNR and SSIM metrics. Moreover, our results look more sharp and close to the target image. Our hyperparameter $\alpha$ in this case is $0.02$.

\begin{table*}
	\caption{Average PSNR(dB) and SSIM for CSGM, GlowIP, MAP, BM3D and our method. We evaluate on inverse problems such as denoising, deblurring, inpainting and colorization. Each numerical measurement is the average over $192$ samples.}
	\label{tab:measures}
	\centering
	\begin{tabular}{@{}lcccccccc@{}}
		\toprule
		Method & \multicolumn{2}{c@{}}{Denoising} & \multicolumn{2}{c@{}}{Deblurring} & \multicolumn{2}{c@{}}{Inpainting} & \multicolumn{2}{c@{}}{Colorization}\\
		\cmidrule(l){2-9}
		& PSNR & SSIM & PSNR & SSIM & PSNR & SSIM & PSNR & SSIM\\
		\midrule
		CSGM   \cite{bora2017compressed} & $37.41$ & $0.897$ & $37.17$ & $0.883$ & $36.39$ & $0.839$ & $34.71$ & $0.897$\\
		GlowIP \cite{asim2020invertible} & $37.21$ & $0.894$ & $35.82$ & $0.867$ & $37.81$ & $0.903$ & $\mathbf{35.79}$ & $0.937$ \\
		MAP \cite{whang2021solving} & $37.35$ & $0.898$ & $-$ & $-$ & $-$ & $-$ & $-$ & $-$ \\
		BM3D   \cite{dabov2007image} & $37.65$ & $0.905$ & $-$ & $-$ & $-$ & $-$ & $-$ & $-$\\
		Ours  & $\mathbf{37.72}$ & $\mathbf{0.918}$ & $\mathbf{39.72}$ & $\mathbf{0.963}$& $\mathbf{37.97}$ & $\mathbf{0.913}$ & $35.72$ & $\mathbf{0.940}$ \\
		\bottomrule
	\end{tabular}
\end{table*}

\subsection{Inpainting}

For inpainting, we mask a square in the center of the target image and fill it with zeros. We obtain the best performance in CSGM for $\gamma=0.01$. But even so, CSGM produces many visual artifacts. In contrast, GlowIP ($\gamma=0$) produces a realistic reconstruction of the black square with few visual artifacts. As we can see in Table \ref{tab:measures}, our method slightly outperforms GlowIP, however our visual results look similar to those from GlowIP (see Figure \ref{fig:inpainting}). Apparently, the log-likelihood term does not significantly improve the performance. We obtain our best performance for $\alpha=0.002$.

\subsection{Colorization}

\begin{figure*}
	\centering
	\hspace{-1.5mm}
	\rotatebox{90}{\hspace{1.8mm}Target}
	\includegraphics[height=1.35cm,trim={0cm 0 15.95cm 0},clip]{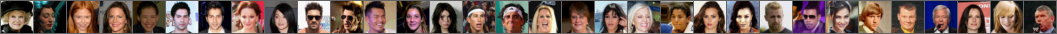}\\
	\hspace{-1.5mm}
	\rotatebox{90}{\hspace{2.8mm}Gray}
	\includegraphics[height=1.35cm,trim={0cm 0 15.95cm 0},clip]{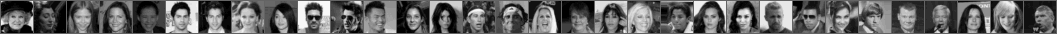}\\
	\rotatebox{90}{\hspace{1.2mm}CSGM}
	\includegraphics[height=1.35cm,trim={0cm 0 15.95cm 0},clip]{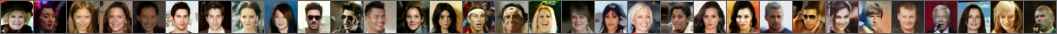}\\
	\rotatebox{90}{\hspace{1mm}GlowIP}
	\includegraphics[height=1.35cm,trim={0cm 0 15.95cm 0},clip]{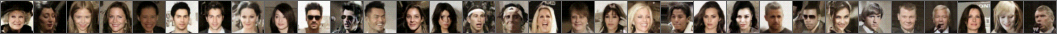}\\
	\rotatebox{90}{\hspace{2.8mm}Ours}
	\includegraphics[height=1.35cm,trim={0cm 0 15.95cm 0},clip]{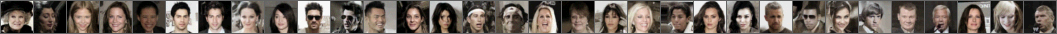}
	
	\caption{Image colorization on images from the CelebA dataset. The first row is the target image $x^{*}$ and the second row is the grayscale image $y$, we aim to recover $x^{*}$ from $y$. From the 3rd-5th rows, we show results for CSGM, GlowIP, and our proposed solver. We create grayscale images by averaging the channels of the target images.}
	\label{fig:colorization}
\end{figure*}

In the case of the colorization problem, we average the three channels of the target image to produce $y$. CSGM ($\gamma=0.01$) generates over-saturated images, which have different tones than target images as we can see in Figure \ref{fig:colorization}. Instead, GlowIP ($\gamma=0$) produces washed-out images, and similar to image deblurring, its regularization term does not improve its performance. Our method performs the colorization problem, however our results are similar to those from GlowIP. In this case, our solver does not achieve a significant improvement over state-of-the-art methods. Table \ref{tab:measures} shows our results for $\alpha=0.02$.

\subsection{Additional experiments}

%Our results show that Equation \ref{eq:1} produces some blur and visual artifacts in many inverse problems such as denoising and deblurring. This leads to unrealistic reconstructions for CSGM and GlowIP in image denoising and image deblurring. Although the $z_0\sim\mathcal{N}(\mathbf{0},\mathbf{I})$ initialization produces saturated images in image colorization, Table \ref{tab:measures} shows better performance for $z_0=\mathbf{0}$. Additionally, we find empirically that for CSGM is convenient to set $\gamma>0$ on its Equation \ref{eq:1} to perform inverse problems. Instead, GlowIP produces its best results with $\gamma=0$ in some inverse problems.

% Additionally, we also show the visual results for image deblurring, image colorization and image inpainting. Figures \ref{fig:deblurring}, \ref{fig:inpainting} and \ref{fig:colorization} show visual results for $9$ samples. For these inverse problems, we compare with CSGM \cite{bora2017compressed} and GlowIP \cite{asim2020invertible}.

Due to our generative flow is trained with a latent distribution $\mathcal{N}(\mathbf{0},\mathbf{I})$, we must use this distribution to produce synthetic samples. Hence, we experiment with different standard deviations of the latent distribution. Given a latent variables $z\sim\mathcal{N}(\mathbf{0},\sigma^2\mathbf{I})$, we sample for many values of $\sigma$ as you can see in Figure \ref{fig:samples}. Our experiments show that if we set $z_0=\mathbf{0}$, the sample $F^{-1}(z_0)=x_0$ is a blurry and unrealistic image. As we increase the standard deviation in the latent variables, we generate high likelihood samples, however we also introduce visual artifacts. Our experiments show that latent variables with a $\sigma$ in $[0.8,1.2]$ produce the most realistic samples.

As you can see in Figure \ref{fig:samples}, high likelihood latent variables do not ensure realistic samples. Both CSGM and GlowIP have problems in image denoising and image deblurring. In the case of inpainting, all methods (including ours) can not perform the reconstruction of non-frontal images. Maybe, a deeper generative flow could generate more realistic restorations---we perform a shallow generative flow for our experiments. An analysis of continuous normalizing flows \cite{chen2018nuralode} as a pre-trained model is beyond the scope of this work.

%------------------------------------------------------------------------

\subsection{Ablation study}

We conduct an ablation study to verify the computational cost on the generation process. As we mentioned before, solving inverse problems implicates using the inverse of our network. Thus, we compare two configurations of our architecture. The first one introduces permutations with invertible $1\times 1$ convolutions; we follow the configuration of Glow~\cite{kingma2018glow} with three different scales ($L=3$). In this configuration, each flow step is a sequence of activation normalization, invertible $1\times 1$ convolution and coupling layer. The second architecture employs our $1\times1$ coupling functions for permutations. Both configurations set $14$ as the total number of flow steps.

Table~\ref{tab:ablation} proves that our method is faster than invertible $1\times1$ convolutions in the  generation process. It is worth mentioning that, on average, our generation time is about $6.4$ms less than the other configuration. Note that both architectures have a similar amount of parameters. Nevertheless, invertible $1\times 1$ convolutions require the computation of its inverse in Equation~\ref{eq:9}. This process increases the time needed to generate a sample.

\begin{table}
	\caption{We compare two architectures with respect to the number of parameters and time speed measured using milliseconds. We configure the first one with invertible $1\times 1$ convolutions, and the second one with our proposed $1\times 1$ coupling functions. The numerical measurements for time speed show the mean and standard deviation over $1500$ iterations and $128$ samples.}
	\label{tab:ablation}
	\centering
	\begin{tabular}{@{}lcc@{}}
		\toprule
		Method & Time (ms) & \# of parameters \\
		\midrule
		Inv. $1\times1$ Conv.   \cite{kingma2018glow} & $10.86\pm 0.18$ & $33.8$M\\
		Ours  & $\mathbf{4.46\pm 0.03}$ & $\mathbf{32.7}$\textbf{M} \\
		\bottomrule
	\end{tabular}
\end{table}

\section{Discussion}

We demonstrate that our solver improves the performance at denoising, deblurring and inpainting in low-resolution images from the CelebA dataset~\cite{liu2015deep}. Indeed, our reconstructions are the sharpest and closest to the original image. Furthermore, we introduce $1\times1$ coupling functions to save computational time in the generation process of generative flows. Our method ensures the generation of high likelihood samples, which results in sharp and realistic restorations in inverse problems. Our experiments suggest that generative flows can be a general purpose solution for inverse problems. Note that we use the same network for each inverse problem, we do not need to re-train a new network for each task. A future direction is to evaluate our method on high-resolution images and real-world examples.% and compare it with state-of-the-art methods for each inverse problem. 

%%%%%%%%% REFERENCES
{\small
	\bibliographystyle{ieee_fullname}
	\bibliography{egbib}
}

%	\appendix

% Thus, in theory, the reconstruction in an inverse problem must have a latent variable with a standard deviation in this range. Indeed, GlowIP increases the standard deviation of the latent variables as we optimize its objective function. In contrast, CSGM needs to regularize the standard deviation of the latent variables because it starts at $z\sim\mathcal{N}(\mathbf{0},\sigma\mathbf{I})$

\end{document}